\documentclass{ecai}
\usepackage{times}
\usepackage{graphicx}
\usepackage{latexsym}

\usepackage{amssymb}
\usepackage{amsfonts}
\usepackage{amsmath}
\usepackage{bm}

\ecaisubmission   

\begin{document}

\title{Gaussian Process Latent Variable Model Factorization\\
 for Context-aware Recommender Systems}

\author{Wei Huang \institute{University of Technology Sydney, Australia, email: Wei.Huang-6@student.uts.edu.au}
    \and Richard Yi Da Xu \institute{University of Technology Sydney, Australia, email: YiDa.Xu@uts.edu.au} }

\maketitle
\bibliographystyle{ecai}

\begin{abstract}
  Context-aware recommender systems (CARS) have gained increasing attention due to their ability to utilize contextual information. Compared to traditional recommender systems, CARS are, in general, able to generate more accurate recommendations. Latent factors approach accounts for a large proportion of CARS. Recently, a non-linear Gaussian Process (GP) based factorization method was proven to outperform the state-of-the-art methods in CARS. Despite its effectiveness, GP model-based methods can suffer from over-fitting and may not be able to determine the impact of each context automatically. In order to address such shortcomings, we propose a Gaussian Process Latent Variable Model Factorization (GPLVMF) method, where we apply an appropriate prior to the original GP model. Our work is primarily inspired by the Gaussian Process Latent Variable Model (GPLVM), which was a non-linear dimensionality reduction method. As a result, we improve the performance on the real datasets significantly as well as capturing the importance of each context. In addition to the general advantages, our method provides two main contributions regarding recommender system settings: (1) addressing the influence of bias by setting a non-zero mean function, and (2) utilizing real-valued contexts by fixing the latent space with real values.  
\end{abstract}

\section{Introduction}
With the advent of the era of big data, users are suffering from the information overload problem. Recommender systems are designed to help users find out items of interest, while context-aware recommender systems (CARS) refine recommendations by exploiting additional contextual information that can have an impact on users' behavior. For example, a user may have preferences on (1) the time of weekday (or weekend) when he/she watches a movie, and (2) the location of office (or home) where he/she uses a mobile phone app. Thus, researchers have introduced CARS that extend the user--item modeling to a more complex user--item--context modeling.

Among existing paradigms for recommender systems, {\it collaborative filtering} approaches that predict the interests of a user by collecting preferences from other users are most popular due to their good accuracy and scalability \cite{koren2015advances}. Because of this reason, it is also the focus of this paper. We concentrate on the class of the latent factor methods based on collaborative filtering that typically involves matrix factorization methods \cite{koren2009matrix} for traditional recommender systems and tensor factorization methods \cite{karatzoglou2010multiverse} for CARS.

In the realm of latent factor methods for CARS, there are two popular classes of approaches. The first class is based on tensor factorization methods \cite{karatzoglou2010multiverse,zheng2016cmptf}, which extend the classical two-dimensional matrix factorization to an n-dimensional tensor factorization. The second class is based on factorization machines \cite{rendle2011fast,blondel2016higher} which originated from a general predictor \cite{rendle2010factorization} that can use second-order feature combinations efficiently. However, both classes adopt the linear combination of second-order or higher-order latent factors to represent user--item--context interactions. In this work, we seek a natural way to capture the inherent non-linear structure of real-world data by the Gaussian process (GP). 

GP is a widely used stochastic process in machine learning \cite{rasmussen2006gaussian}. Standard GP models can only deal with supervised machine learning tasks while an approach called Gaussian process latent variable model (GPLVM) \cite{lawrence2004gaussian} is designed to address un-supervised problems. Supervised GP learning of user preferences and un-supervised dimensionality reduction from ratings to latent factors constitute the foundation of GP-based collaborative factorization methods. Two works that investigate non-linear matrix factorization for conventional recommender systems \cite{lawrence2009non} and non-linear factorization for CARS \cite{nguyen2014gaussian} are proposed. Both of them use the Gaussian process to realize non-linear factorization and have gained state-of-the-art performances compared to linear methods. 

Despite the successful application of the Gaussian process latent variable model to context-aware recommender systems, the model itself cannot infer the impact of each context and is sensitive to overfitting because it does not marginalize out the latent variables \cite{damianou2015deep}. To address these two drawbacks of GP-based factorization for CARS, we introduce a prior to the latent variables. Inspired by Gaussian process latent variable model \cite{titsias2010bayesian}, where a variational inference framework for training is applied, we further implement a generalized variational inference that includes a non-zero mean function during the GP to solve the model. Besides, our model is flexible enough to integrate both categorical and real-valued contexts by fixing the latent variable with corresponding real values for real-valued contexts while adding a prior to the latent variable for categorical contexts.

As a result, we have developed a powerful non-linear collaborative filtering method, which we name, Gaussian Process Latent Variable Model Factorization (GPLVMF), to improve the performance of GP-based factorization methods for CARS further . To summarize, the main contributions of this paper are:

\begin{itemize}
\item We propose a novel algorithm named GPLVMF to achieve a Bayesian non-linear factorization for CARS. Different from GP-based methods, GPLVMF aims to address the over-fitting problem via implementing a  prior distribution to the latent variables.
\item We model the non-zero mean function during Gaussian processes to capture the bias and provide a generalized variational inference solution. Also, our method can flexibly deal with both real-valued and categorical contexts.
\item Both scaled conjugate gradient (SCG) and stochastic gradient descent (SGD) optimization methods are applied to solve the model. The experiment results show that GPLVMF not only improves the accuracy of the real datasets but also can automatically infer the influence of each context.
\end{itemize}

\section{Related Work}

There are three main approaches to integrate context information into recommender systems \cite{adomavicius2011}, namely: (1) {\it contextual pre-filter} where contextual information is used as a filter before applying context \cite{adomavicius2005incorporating,otebolaku2015context}, (2) {\it contextual post-filtering} where contextual information is initially ignored and is then filtered after using traditional recommender algorithms \cite{ramirez2014post}, and (3) {\it contextual modeling} where contextual information is integrated into the process of modeling directly, which is what this paper is focusing on, since it does not require supervision and fine-tuning in all steps \cite{rendle2011fast}.  
	
In {\it contexture modeling}, we introduce three classes of approaches. The first class is based on the matrix factorization \cite{baltrunas2011matrix}. Later, \cite{liu2015learning} used biased matrix factorization as the base model while \cite{yu2017low} adopted matrix completion modeling to address CARS. The second class is based on tensor factorization \cite{karatzoglou2010multiverse}. Then \cite{hidasi2012fast,shi2012tfmap} proposed tensor factorization-based methods for implicit feedback data. The final one is based on factorization machines \cite{rendle2010factorization}. A factorization machines-based method to solve the cross-domain problem was proposed by \cite{loni2014cross} while a higher order factorization machine was proposed by \cite{blondel2016higher}. However, these methods treat user--item--context interactions as linear combinations of latent factors and are insufficient to capture the complex non-linear inter-relationships between the three entities in CARS.

GP has been applied to the recommendation problems. A non-linear method for matrix factorization based on GP \cite{lawrence2009non} for traditional recommender systems was shown to outperform standard matrix factorization methods. Collaborative Gaussian processes for preference learning \cite{houlsby2012collaborative} was proposed to learn pairwise preferences expressed by multiple users. Besides, the GP was used to address the challenge of ranking recommendation on click feedback recommender system \cite{vanchinathan2014explore}. Recently, a GP-based factorization machine for context-aware recommender systems was proposed \cite{nguyen2014gaussian}, and it can outperform factorization machines and tensor factorization methods. However, none of these methods have studied the case of applying Gaussian process latent variable model to the CARS. 


\section{Gaussian Process Latent Variable Model Factorization}

\begin{figure*}[htbp]
		\begin{center}   
			\includegraphics[width=0.85\linewidth]{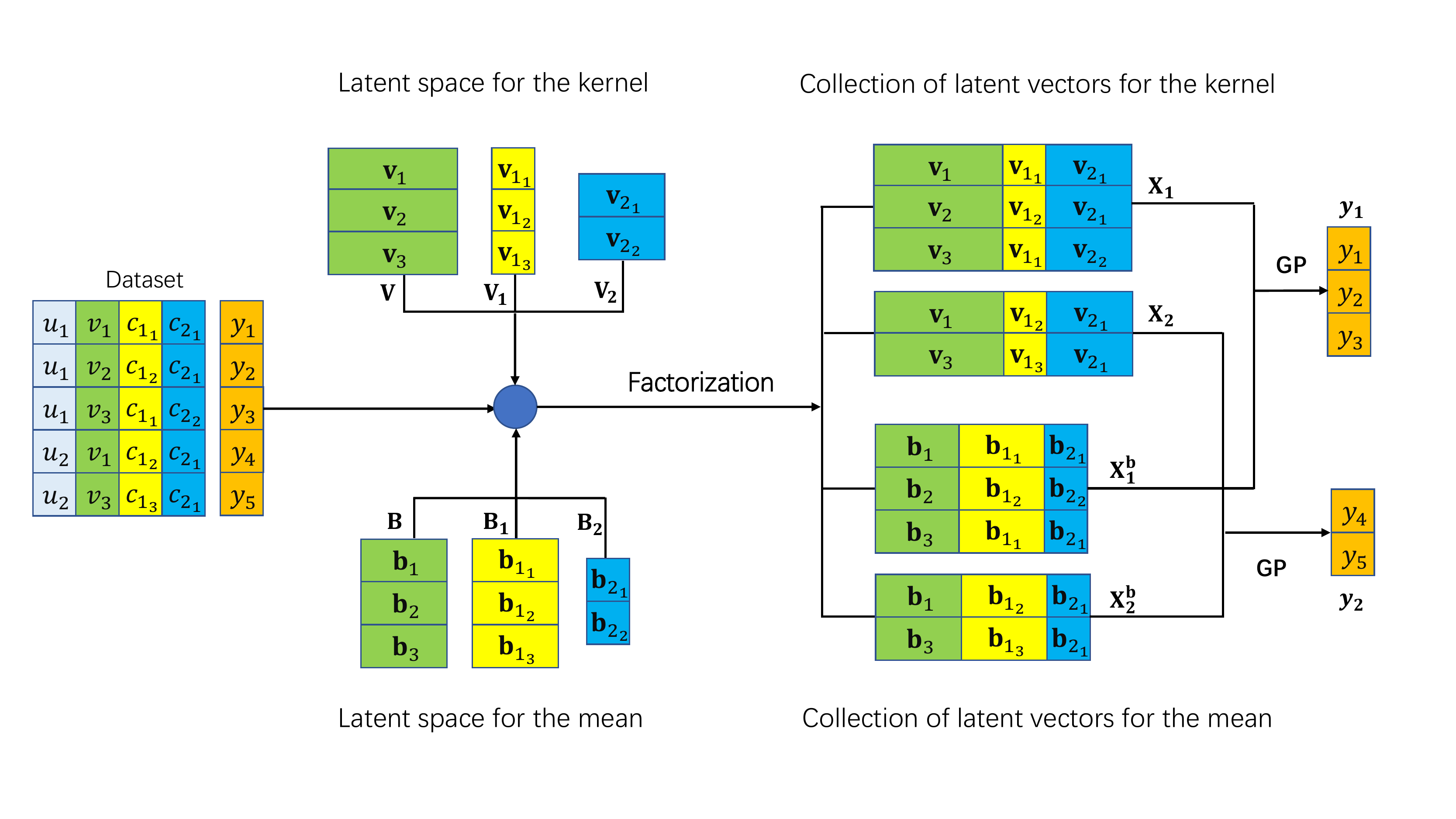}
			\caption{Overview of our proposed GPLVMF model. First, the recommendation dataset consisted of two users is shown on the left side. Grey is the user, green is the item, yellow is the first context, blue is the second context, and orange is the rating. The latent space for the mean and kernel are in the middle part, the width of each rectangle represents the dimensionality of each latent vector. We use the identical color and subscript as the items, first contexts, and second contexts in the dataset to indicate the corresponding latent vectors. Finally, GPs from the collection of latent vectors to ratings are illustrated on the right side. The latent space is factorized into two groups for both the mean and kernel according to two users. \label{overview}}
		\end{center}
	\end{figure*}
	
In this section, we elaborate on our proposed Bayesian Gaussian process factorization method. First, we explain how to use latent factors to represent the user--item--context interactions for CARS. Then we describe the Gaussian process latent variable model factorization via introducing a prior to latent factors. Finally, we derive a variational inference for solving the model.

\subsection{Latent Representation}
	
A conventional recommender system is an information filtering system that seeks to predict the rating a user would give to an item. We denote users and items by $U = \{u_n\}_{n=1}^N$ and $V = \{v_l\}_{l=1}^L$ respectively, where $N = |U|$ and $ L = |V|$. When it comes to the context-aware setting, multiple contexts can be denoted by $\mathcal{C}_1, \ldots, \mathcal{C}_D$ to describe their contextual information (location, time etc.), where $ \mathcal{C}_d = \{ c_{d_l} \}_{l=1}^{L_d}$, and $L_d = |\mathcal{C}_d|$, for each $d \in [1,D]$. 
	
Latent factors approach defines transformation from observation to its latent representation. Let each element of the entities, including user, item, and each of the multiple contexts be represented by a real-valued vector in the latent space $\textbf{U} =\{ \textbf{u}_n \}_{n=1}^N $, $\textbf{V} =\{ \textbf{v}_l \}_{l=1}^L$, and $\textbf{V}_d = \{ \textbf{v}_{d_l} \}_{l=1}^{L_d}$ for each $d \in [1,D]$. Suppose the dimension of latent spaces for users, items, and contexts are $Q_u$, $Q_v$, and $\{Q_d\}_{d=1}^D$ respectively. Then, we have $\textbf{U} \in \mathbb{R}^{N \times Q_u} $, $\textbf{V} \in \mathbb{R}^{L \times Q_v}$ and $\textbf{V}_{d} \in \mathbb{R}^{L_d \times Q_d} $. Here, we use bold capital letters to denote matrices, bold letters for vectors, and regular letters for scalars. 

\subsection{Gaussian Process}

A Gaussian process prior defines a distribution over a real-valued function $f(\textbf{x})$. Formally, the  set of function values $ \textbf{f} =\{f(\textbf{x}_t)\}_{t=1}^{T}$ on a collection of any finite inputs $\textbf{X} = \{\textbf{x}_t\}_{t=1}^T$ should satisfy the multivariate Gaussian distribution. A GP is completely specified by its mean function $m(\textbf{x})$ and covariance function $k(\textbf{x}, \textbf{x}')$, which denotes as $f(\textbf{x}) \sim \mathcal{GP}(m(\textbf{x}),k(\textbf{x},\textbf{x}'))$, and distribution can be written as: 
    \begin{equation}\label{eq:gaussian}
    p(\textbf{f}| \textbf{X})= \mathcal{N}(\textbf{f}|\textbf{m},\textbf{K}),
    \end{equation}
where $\textbf{m} = \{ m(\textbf{x}_t)\}_{t=1}^T $ is a vector and $\textbf{K}$ is a covariance matrix where each element corresponds to the value of covariance function evaluated between all pairs of $\textbf{x}_t, \textbf{x}_{t'}$. For common usage of GP, the mean function is usually assumed to be zero, $m(\textbf{x}) = 0$, which is not the case in our work. Different covariance functions will be applied to different applications while a popular example is the RBF kernel,
    \begin{equation}
    k(\textbf{x}, \textbf{x}')= \sigma^2 \exp (-\frac{1}{2\ell^2 }(\textbf{x}-\textbf{x}')^T(\textbf{x}-\textbf{x}')), 
    \end{equation}
where $\sigma$ is known as the {\it signal variance} and $\ell$ the {\it length-scale}. A Gaussian likelihood function between observations $\textbf{y} = \{y_t\}_{t=1}^{T}$ and prior $\textbf{f}$ accounts for the noise, 
    \begin{equation}\label{eq:noise}
     p(\textbf{y}| \textbf{f})= \mathcal{N}(\textbf{y}|\textbf{f},\beta ^{-1}\textbf{I}),
    \end{equation} 
where $\beta$ is the {\it noise precision} and $\textbf{I}$ is an identity matrix with size $T$. Integrating out vector functions $\textbf{f}$, we can obtain the marginal likelihood function,
    \begin{equation}\label{eq:observation}
    p(\textbf{y}| \textbf{X}) = \int p(\textbf{y}| \textbf{f})p(\textbf{f}| \textbf{X})d\textbf{f}
                             = \mathcal{N}(\textbf{y}|\textbf{m},\textbf{K} + \beta ^{-1}\textbf{I}).
    \end{equation}

\subsection{Mean and Kernel for CARS}

For standard GP regression and classification tasks, the mean function for bias is usually assumed to be zero, since data bias can be eliminated by pre-procession. However, in recommendation system bias is a well-known phenomenon. For example, some users may have their preferences on certain items and consistently give high ratings to them. To mitigate this problem, we allocate additional variables to users, items, and contexts. Let $\textbf{B} =\{ \textbf{b}_l \}_{l=1}^L$, and $\textbf{B}_d = \{ \textbf{b}_{d_l} \}_{l=1}^{L_d}$ for each $d \in [1,D]$ represent latent space of corresponding entities for the mean, and $Q^b_v$ and $Q^b_d$ are the dimension of corresponding latent space.

For each rating associated with user $n$ to item $l$ under contexts $d_l$, we define a mean function:
    \begin{equation}\label{eq:bias}
    m_n(\textbf{x}^b) = b_n + \sum_{q=1}^{Q^b_v} \textbf{b}_{l,q} + \sum_{d=1}^D \sum_{q=1}^{Q^b_d} \textbf{b}_{d_l,q} ,
    \end{equation}    
where $b_n$ is a parameter regarding the bias of user $n$ and $\textbf{x}^b$ represents a latent vector that consists of $\textbf{b}_l, \textbf{b}_{d_l}$, for each $d \in [1,D]$. 

Besides, we focus on the automatic relevance determination (ARD) squared exponential kernel between latent vectors associated with two ratings for user $n$:
\begin{equation}\label{eq:kernel}
  \begin{aligned} 
  k_n(\textbf{x}, \textbf{x}')=  \sigma_n^2 & \exp \Bigg(  -\frac{1}{2}\sum_{q=1}^{Q_v}\alpha_q (\textbf{v}_{l,q}-\textbf{v}_{l',q})^2 \\
  &-\frac{1}{2}\sum_{d=1}^D \sum_{q=1}^{Q_d}\alpha_q(\textbf{v}_{d_l,q}-\textbf{v}_{d_{l'},q})^2 \Bigg ) ,
  \end{aligned} 
\end{equation} 
where $\alpha_q $ is {\it inverse length-scale} and $\textbf{x}$ represents a latent vector that consists of $\textbf{v}_m, \textbf{v}_{d_z}$ for each $d \in [1,D]$. ARD squared exponential kernel can help the system to automatically select the dimensionality of its latent features \cite{titsias2010bayesian}. Therefore, these weights $\alpha_q$ can give us the insights into the impact of each context.

\subsection{Probabilistic Factorization}

After introducing Gaussian process, we seek a natural probabilistic interpretation of factorization based on GPLVM. We show the overview of GPLVMF in Figure \ref{overview}. Let $ \textbf{X}_n \in \mathbb{R}^{N_n \times Q}$ and  $ \textbf{X}^b_n \in \mathbb{R}^{N_n \times Q^b}$ denote the collection of latent variables regarding rated entities by user $n$ for the kernel and mean respectively, where $N_n$ is the number of ratings by user $n$, $Q = Q_v +\sum_{d=1}^D Q_d$, and $Q^b = Q^b_v +\sum_{d=1}^D Q^b_d$. We use $\textbf{y}_n \in \mathbb{R}^{N_n}$ to denote the corresponding ratings by user $n$. Our user-centric factorization method for the context-aware recommender systems takes the probabilistic form:
     \begin{equation}\label{eq:likelihood}
     P(\textbf{Y}\vert\textbf{X})= \prod_{n=1}^N p(\textbf{y}_n \vert \textbf{X}_n,\textbf{X}^b_n) ,
     \end{equation}
where $\textbf{X} $ is the collection of all the latent vectors, and $\textbf{Y} = \{\textbf{y}_n\}_{n=1}^N$. GPs are taken to be independent across of different users and the likelihood function for user $n$ is written as:
     \begin{equation}\label{eq:likelihood_user}
     p(\textbf{y}_n \vert \textbf{X}_n,\textbf{X}^b_n,\bm{\alpha},b_n,\theta_n) = \mathcal{N}(\textbf{y}_n \vert \textbf{m}_n, \textbf{K}_n + \beta_n^{-1}\textbf{I}_{N_n}) ,
     \end{equation}     
where $\bm{\alpha} = \{\alpha_q\}_{q=1}^Q$ and $\theta_n = (\sigma_n, \beta_n)$. We'd like to mention that the latent space of users are marginalized out in the likelihood function, while we could integrate latent space of items instead to obtain another likelihood function as well \cite{lawrence2009non}. 
        
The optimization method adopted by GP factorization for both the conventional \cite{lawrence2009non} and context-aware \cite{nguyen2014gaussian} recommender system is to find the maximum likelihood estimation of latent variables $\textbf{X}$ whilst jointly maximizing with respect to the hyper-parameters $ (\bm{\alpha},\theta_n)$. However, this method is sensitive to over-fitting and cannot determine the dimensionality of latent space automatically for GPLVM \cite{titsias2010bayesian}.

To adopt a fully Bayesian treatment for the latent space, we assign to it, a prior density over $\textbf{X}$. In this work, we use the standard normal density for the latent variables while we use a $\delta$ function prior to realize fixing the latent variables regarding real-valued contexts. The normal distribution for the $\textbf{X}$ can be written as:
     \begin{equation}\label{eq:prior}
     p(\textbf{X}) = \prod_{l=1}^{L}\mathcal{N}(\textbf{v}_l \vert \textbf{0}, \textbf{I}_{Q_v}) \prod_{d=1}^{D}\prod_{l=1}^{L_d} \mathcal{N}(\textbf{v}_{d_l} \vert \textbf{0}, \textbf{I}_{Q_d}).
     \end{equation}
We note the distribution of the latent variables for the mean function has the same form as the kernel function and is dropped from the expression for simplification. The joint probability for the model is: 
     \begin{equation}\label{eq:joint}
     p(\textbf{Y},\textbf{X}) = \prod_{n=1}^N p(\textbf{y}_n\vert\textbf{X}_n,\textbf{X}_n^b)p(\textbf{X}).
     \end{equation}
However, this is not analytically tractable. While a sampling method was proposed by \cite{salakhutdinov2008bayesian} for Bayesian probabilistic matrix factorization problem, a variational inference approach to marginalize the latent variables was proposed by \cite{titsias2010bayesian}. Inspired by the latter work, we derive a generalized variational inference for Gaussian process latent variable model factorization to include the non-zero mean Equation (\ref{eq:bias}).

\subsection{Variational Inference}

We aim  to compute the marginal data likelihood:
    \begin{equation}\label{eq:marginal}
    p(\textbf{Y}) = \int p(\textbf{Y} \vert \textbf{X})p(\textbf{X}) d\textbf{X} .
    \end{equation}
However, the integral of Equation (\ref{eq:marginal}) is intractable due to the non-linearity inside the inverse of the covariance matrix. Following the VI framework, we introduce a variational distribution $q(\textbf{X})$ to approximate the true posterior $p(\textbf{X}\vert\textbf{Y})$, 
    \begin{equation}\label{eq:variational prior}
    q(\textbf{X}) = 
    \prod_{l=1}^{L}\mathcal{N}(\textbf{v}_l \vert \bm{\mu}_l, \textbf{S}_l) \prod_{d=1}^{D}\prod_{l=1}^{L_d} \mathcal{N}(\textbf{v}_{d_l} \vert \bm{\mu}_{d_l}, \textbf{S}_{d_l}),
    \end{equation}
where the variational parameters for items and contexts are $ \bm{\mu} = ( \{ \bm{\mu}_l \}, \{ \bm{\mu}_{d_l} \})$ and $ \bm{S} = ( \{ \bm{S}_l \}, \{ \bm{S}_{d_l} \})$. Again, we drop the variables regarding the mean from expressions for simplification. We apply Jensen's inequality to find a lower bound $ F(q) \leqslant \log p(\textbf{Y})$, and the lower bound $F(q)$ takes the form:
	\begin{equation}\label{eq:lower bound}
	\begin{aligned}
    F(q) &  = \int q(\textbf{X}) \log  \frac{p(\textbf{Y}\vert\textbf{X})p(\textbf{X})}{q(\textbf{X})} d             \textbf{X} \\ 
    & = \int q(\textbf{X}) \log p(\textbf{Y}\vert\textbf{X})d\textbf{X} - \int q(\textbf{X})\log \frac{q(\textbf{X})}{p(\textbf{X})} d\textbf{X}\\
    & = \tilde{F}(q) -{\rm KL} (q||p),
    \end{aligned}
    \end{equation}
    where $\rm{KL}$ denotes the Kullback -- Leibler divergence, which can be computed analytically thanks to the fact that distributions of $q(\textbf{X})$ and $p(\textbf{X})$ are both Gaussians. The first term can further be broken down to separate form for each user,
    \begin{equation}\label{eq:tilde lower bound}
    \begin{aligned}
    \tilde{F}(q)  = \sum_{n=1}^N \tilde{F}_n(q) & = \sum_{n=1}^N \int q(\textbf{X})\log p(\textbf{y}_n \vert \textbf{X}_n) d\textbf{X} \\ 
         & = \sum_{n=1}^N \int q(\textbf{X}_n)\log p(\textbf{y}_n \vert \textbf{X}_n) d\textbf{X}_n. \\
    \end{aligned}
    \end{equation}
Note that for CARS, $\textbf{X}_n \subseteq \textbf{X}$, so that the complementary latent variables can be integrated out. 

To further solve the Equation (\ref{eq:tilde lower bound}), we need to introduce a variational spare Gaussian process framework \cite{titsias2010bayesian} to modify Gaussian process prior. For each vector of latent function values $\textbf{f}_n$, we introduce a separate set of $M$ inducing variables $\textbf{u}_n \in \mathbb{R}^M$ evaluated at a set of inducing input locations given by $\textbf{Z} \in \mathbb{R}^{ M \times Q}$. The likelihood associated with GP latent function $p(\textbf{y}_n, \textbf{f}_n \vert \textbf{X}_n) = p(\textbf{y}_n\vert \textbf{f}_n) p(\textbf{f}_n \vert \textbf{X}_n)$, 
is augmented by inducing variables:
	\begin{equation}\label{eq:likelihood_fu}
    p(\textbf{y}_n, \textbf{f}_n, \textbf{u}_n \vert \textbf{X}_n, \textbf{Z}) = p(\textbf{y}_n\vert \textbf{f}_n) p(\textbf{f}_n \vert \textbf{u}_n, \textbf{X}_n, \textbf{Z}) p(\textbf{u}_n \vert \textbf{Z}).
    \end{equation} 
Both $\textbf{f}_n$ and $\textbf{u}_n$ are from same distribution, thus $p(\textbf{f}_n \vert \textbf{u}_n, \textbf{X}_n, \textbf{Z}) = \mathcal{N}(\textbf{f}_n \vert \bm{\alpha}_n,\bm{\Sigma}_n)$, where
   \begin{equation}\label{eq:alpha}
   \begin{aligned}
 	&\bm{\alpha}_n = \textbf{m}_n + \textbf{K}_{N_n M} \textbf{K}^{-1}_{MM} \textbf{u}_n, \\
   &\bm{\Sigma}_n = \textbf{K}_{N_n N_n} - \textbf{K}_{N_n M} \textbf{K}^{-1}_{MM}\textbf{K}_{M N_n} .
   \end{aligned}
    \end{equation} 
Note we include the bias term $\textbf{m}_n$ in $\bm{\alpha}_n$, which is the key of generalized form for VI. The next step is to adopt a variational distribution:
	\begin{equation}\label{eq:variational_fu}
    q(\textbf{f}_n, \textbf{u}_n) = p(\textbf{f}_n \vert \textbf{u}_n, \textbf{X}_n) q(\textbf{u}_n),
    \end{equation}
where $q(\textbf{u}_n)$ is a variational distribution over the inducing variables $\textbf{u}_n$. Here we simplify our notation by dropping $\textbf{Z}$ from our expressions. By Jensen's inequality, we obtain a lower bound for the log-likelihood:
    \begin{equation}\label{likelihood lower bound}
	\begin{aligned}
    \log p(\textbf{y}_n \vert \textbf{X}_n) &  \geqslant \int q(\textbf{u}_n) \log  \frac{p(\textbf{u}_n) \mathcal{N}(\textbf{y}_n \vert \bm{\alpha}_n,\beta_n^{-1} \textbf{I}_{N_n})} {q(\textbf{u}_n)} d \textbf{u}_n \\ 
    & - \frac{\beta}{2} {\rm Tr} (\textbf{K}_{N_n N_n} - \textbf{K}_{N_n M} \textbf{K}^{-1}_{MM}\textbf{K}_{M N_n}). \\
    \end{aligned}
    \end{equation}
Substituting Equation (\ref{likelihood lower bound}) back to Equation (\ref{eq:tilde lower bound}) and swapping the integration order, we have:
    \begin{equation}\label{intx lower bound}
	\begin{aligned}
    & \tilde{F}_n(q)  \geqslant \\
    & \int q(\textbf{u}_n) \left[ \langle \log \mathcal{N}(\textbf{y}_n \vert \bm{\alpha}_n, \beta_n^{-1} \textbf{I}_{N_n}) \rangle + \log\frac{p(\textbf{u}_n)} {q(\textbf{u}_n)} \right] d \textbf{u}_n \\ 
    & - \frac{\beta_n}{2} {\rm Tr} (\langle \textbf{K}_{N_n N_n} \rangle) + \frac{\beta_n}{2} {\rm Tr}  (\textbf{K}^{-1}_{MM} \langle \textbf{K}_{M N_n} \textbf{K}_{N_n M} \rangle ),
    \end{aligned}
    \end{equation}
where $\langle \cdot \rangle$ denotes expectation under the distribution $q(\textbf{X})$. 

The last step is to adopt $q(\textbf{u}_n)$ as a variational distribution to maximize the above lower bound \cite{titsias2010bayesian}. The optimal setting of distribution is 
   \begin{equation}\label{setting}
   q(\textbf{u}_n) \propto e^{\langle \log \mathcal{N}(\textbf{y}_n \vert \bm{\alpha}_n, \beta_n^{-1} \textbf{I}_{N_n}) \rangle} p(\textbf{u}_n), 
   \end{equation}
which is a Gaussian distribution $q(\textbf{u}_n) = \mathcal{N}(\textbf{u}_n \vert \bm{\mu}_{u_n}, \bm{\Sigma}_{u_n})$. By calculating, we obtain its mean and covariance:
   \begin{equation}\label{q(ud) gaussin}
	\begin{aligned}
    & \bm{\mu}_{u_n} = \textbf{K}_{MM}(\beta_n^{-1}\textbf{K}_{MM}+ \bm{\Psi}_{2_n} )^{-1}\bm{\Psi}_{1_n}^{\mathrm{T}}(\textbf{y}_n-\bm{\phi}_{1_n}) , \\
    & \bm{\Sigma}_{u_n} =\beta_n^{-1} \textbf{K}_{MM}(\beta_n^{-1}\textbf{K}_{MM}+ \bm{\Psi}_{2_n})^{-1}\textbf{K}_{MM} , \\
    \end{aligned}
    \end{equation}
where $\bm{\Psi}_{1_n} = \langle \textbf{K}_{N_nM} \rangle_{q(\textbf{X}_n)}$, $\bm{\Psi}_{2_n}=\langle \textbf{K}_{MN_n} \textbf{K}_{N_nM} \rangle_{q(\textbf{X}_n)}$ and $\bm{\phi}_{1_n} = \langle \textbf{m}_n \rangle_{q(\textbf{X}^b_n)}$. In the following computation, we would use another two statistics: $\psi_{0_n} = {\rm Tr}(\langle \textbf{K}_{NN} \rangle_{q(\textbf{X}_n)})$ and $\phi_{0_n} = \langle \textbf{m}_n^{\mathrm{T}} \textbf{m}_n\rangle_{q(\textbf{X}^b_n)}$. 
   

Our contribution to the generalized variational inference form is introducing the statistics $ \phi = (\phi_{0}, \bm{\phi}_{1}$) with regard to the mean, while the original statistics $ \Psi = (\psi_{0}, \bm{\Psi}_{1},\bm{\Psi}_{2} $) are accounting for the kernel. Since all terms are tractable now, we finally get the closed-form of the lower bound for $\tilde{F}_n(q)$,
	\begin{equation}\label{final bound}
	\begin{aligned}
    \tilde{F}_n(q) & \geqslant  \log  \Bigg[ \frac{\beta_n^{\frac{N_n}{2}} | \textbf{K}_{MM} |^\frac{1}{2} }{(2\pi)^{\frac{N_n}{2}}|\beta_n \bm{\Psi}_{2_n} + \textbf{K}_{MM}|^{\frac{1}{2}}} e^{-\frac{1}{2} (\textbf{W}_1-\textbf{W}_2)} \Bigg] \\    & - \frac{\beta_n \psi_{0_n}}{2} + \frac{\beta_n}{2} {\rm Tr} (\textbf{K}^{-1}_{MM} \bm{\Psi}_{2_n}) , 
    \end{aligned}
    \end{equation}
    where $\textbf{W}_1 =  \beta_n ( \textbf{y}_n^{\mathrm{T}} \textbf{y}_n-2 \textbf{y}_n ^{\mathrm{T}} \bm{\phi}_{1_n} + \phi_{0_n})$ and $\textbf{W}_2 = \beta_n^2  \tilde{\textbf{y}}_n^{\mathrm{T}}\bm{\Psi}_{1_n}(\textbf{K}_{MM}+\beta_n \bm{\Psi}_{2_n})^{-1} \bm{\Psi}_{1_n}^{\mathrm{T}} \tilde{\textbf{y}}_n$,  here $\tilde{\textbf{y}}_n = \textbf{y}_n-\bm{\phi}_{1_n} $. The bound can be jointly maximized over the variational parameters $( \bm{\mu}, \bm{\mu^b}, \bm{S}, \bm{S}^b, \textbf{Z})$ and model parameters $(\bm{\alpha}, \{b_n,\theta_n\}_{n=1}^{N})$ by applying gradient-based optimization techniques.

The bound can be jointly maximized over the variational parameters $( \bm{\mu}, \bm{\mu^b}, \bm{S}, \bm{S}^b, \textbf{Z})$ and model parameters $(\bm{\alpha}, \{b_n,\theta_n\}_{n=1}^{N})$ by applying gradient-based optimization techniques.

\subsection{Optimization and complexity analysis}

In this work, we apply stochastic gradient descent (SGD) to solve the optimization problem. We note that the gradient of the KL divergence of each step is averaged over the number of users since there is no independent form for the users. In addition, we adopt the scaled conjugate gradient (SCG) optimization \cite{moller1993scaled} to compare with SGD. To counter one epoch of SGD, we use an iteration that processes the whole data once for SCG.

Bayesian Gaussian process latent variable model cannot handle big data problem, since the computational complexity of $\mathcal{O}(NM^2)$ and storage demands of $\mathcal{O}(NM)$ for general application, where $N$ is the size of data and $M$ is the number of inducing point \cite{hensman2013gaussian}. However, we would not encounter this challenge for GPLVMF thanks to the structure of the dataset in the process of factorization, as shown in Figure \ref{overview}. The maximum kernel size is decided by the user who has the most ratings, which is usually no more than millions or billions in the real-world dataset. Finally, the computational complexity and storage demands in the GPLVMF are $ \mathcal{O}(\sum_{n=1}^N N_n M^2)$ and $ \mathcal{O}(\sum_{n=1}^N N_n M)$ at an iteration respectively. SGD would be a better choice when the number of users becomes millions or billions. A complete comparison between SCG and SGD will be presented in the next section. 

\subsection{Prediction}    
   
Once all the variational parameters and model parameters have been learned, they can be used to predict preferences for users. We denote the collection of latent variables for user $n$ for prediction as $\textbf{X}^\ast_n$. Then we have the predictive function $\textbf{f}^\ast_n$,
	\begin{equation}\label{eq:prediction}
    q(\textbf{f}^\ast_n) = \int \bigg(  p(\textbf{f}^\ast_n \vert \textbf{u}_n, \textbf{X}^\ast_n) q(\textbf{u}_n)d\textbf{u}_n  \bigg) q(\textbf{X}^\ast_n) d\textbf{X}^\ast_n,
    \end{equation}
	which is a Gaussian distribution $ \mathcal{N}(\textbf{f}^\ast_n \vert \bm{\mu}^\ast_{f_n}, \bm{\Sigma}^\ast_{f_n}) $ with its mean and covariance:  
	\begin{equation}\label{eq:prediction distribution}
	\begin{aligned}
    & \bm{\mu}^\ast_{f_n} = \bm{\Psi}_{1_n}^\ast(\beta_n^{-1}\textbf{K}_{MM}+\bm{\Psi}_{2_n})^{-1}\bm{\Psi}_{1_n}^{\mathrm{T}}\tilde{\textbf{y}}_n +\bm{\phi}^\ast_{1_n} ,\\
    & \bm{\Sigma}^\ast_{f_n} = \bm{\Psi}_{1_n}^\ast(\textbf{K}^{-1}_{MM}+(\textbf{K}_{MM}+\beta_n \bm{\Psi}_{2_n})^{-1})(\bm{\Psi}_{1_n}^\ast)^{\mathrm{T}},
    \end{aligned}
    \end{equation}
where $\bm{\Psi}^\ast_{1_n} = \langle \textbf{K}_{N_n^\ast M} \rangle_{q(\textbf{X}^\ast_n)}$ and $\bm{\phi}^\ast_{1_n} = \langle \textbf{m}_n^\ast  \rangle_{q(\textbf{X}^{b \ast}_n)}$.

\section{Experiment}


In this section, we empirically investigate the performance of GPLVMF. First we describe the datasets and settings in our experiments, then report and analyze the experiment results.

\subsection{Dataset}

In this work, we use 4 real datasets. The statistics of all datasets are presented in Table \ref{tab:dataset}.
\begin{itemize}
\item Comoda \cite{kovsir2011database} contains 2296 ratings of 1232 movies by 121 users. We use 12 provided contexts: time of the day, day type, season, location, weather, social, ending emotions and dominant emotions, mood, physical conditions, decision, and interaction.
	
\item Food \cite{ono2009context} contains 5554 ratings by 212 users on 20 food menus. We use 2 contexts: three different levels of hunger and real or supposed situation. We have eliminated some conflicted ratings followed by \cite{nguyen2014gaussian}.
	
\item Sushi \cite{kamishima2003nantonac} contains 50,000 ratings of 100 types of Sushi by 5000 Japanese users. We use 7 contexts: style, major group, minor group, heaviness/oiliness in the state, popularity, price, and availability in shops. All the contextual information are attributions of the item, and the last four contexts are real-valued contexts.

\item Movielens-1M \cite{harper2016movielens} contains 1,000,209 ratings of 3706 movies by 6040 users. In this work, we adopt the hour and day as 2 contexts.
\end{itemize}

    \begin{table}
		\caption{Statistics of Real Datasets}
		\label{tab:dataset}
		\begin{center}
			\label{st}
			\begin{tabular}{|c|c|c|c|c|c|c|}
				\hline
				dataset  & user & item & context & observables & rating\\
				\hline
				Comoda   & 121   & 1232  & 12        &2296         & 1-5 \\
				Food     & 212   & 20    & 2         &5554         & 1-5 \\
				Sushi    & 5000  & 100   & 7         &50000        & 0-4 \\
         Movielens-1M  & 6040  & 3706  & 2         &1000209      & 1-5 \\
				\hline
			\end{tabular}
		\end{center}
	\end{table}

\subsection{Evaluation} 

We compare our results with state-of-the-art methods, namely,
\begin{itemize}

\item Const, a naive predictor that predicts for every user the mean of his/her ratings. 

\item Multiverse \cite{karatzoglou2010multiverse}, a state-of-the-art tensor factorization method.

\item FM \cite{rendle2011fast}, one of the most popular factorization method, which is famous for its fast training speed. We use LibFM\footnote{http://www.libfm.org/} to implement the method.
	
\item GPFM \cite{nguyen2014gaussian}, a Gaussian process based factorization method and has been shown to outperform other context-aware recommendation models on the Comoda dataset, the Food dataset, and the Sushi dataset. 
	
\item GPLVM--MF, a Gaussian process latent variable model based matrix factorization method. We apply our model to a setting where no contextual information is available. Thus GPLVM--MF can be seen as a context-agnostic variant of GPLVMF. 

\end{itemize}

For each dataset, we split it 5 folds and repeat the experiments 5 times using 1 fold as the test set and the remaining 4 folds as the training set. We tune the parameters using 1 of the 5 folds as the validation set and fix the tunes parameters for other 4 folds. To evaluate the prediction of ratings, we use two evaluations metrics: mean absolute error (MAE), root-mean-square error (RMSE). For all datasets, the performance is averaged over the 5 different folds. The results are statistically significant and the variances are small so not reported.

\subsection{Performance comparison}
	
The performance comparison of all methods are shown in Table~\ref{tab:performance} in terms of MAE and RMSE. This table shows our approach achieves the best results on all datasets, which demonstrates the effectiveness of using Gaussian process latent variable model factorization to model contextual-aware recommendation.

    \begin{table*}[!htbp]
		\caption{Performance comparison on the 4 real datasets in terms of MAE and RMSE}
		\label{tab:performance}
		\begin{center}
			\begin{tabular}{|c |c c| c c|c c| c c|c c|}
			\hline
			Dataset  & \multicolumn{2}{c|}{Comoda}  & \multicolumn{2}{c|}{Food}  & \multicolumn{2}{c|}{Sushi}  & \multicolumn{2}{c|}{Movielens-1M} \\
			\hline 
			Performance  & MAE & RMSE  & MAE & RMSE  & MAE & RMSE  & MAE & RMSE  \\
			\hline
		   GPLVMF     &\textbf{0.6708} &\textbf{0.8548}  &\textbf{0.6328} &\textbf{0.8356}  &\textbf{0.9006} &\textbf{1.1633}  &\textbf{0.6701}  &\textbf{0.8543}  \\ 
		   GPFM       &0.6991 &0.8879  &0.7311 &0.9580 &0.9111 &1.1643  &0.7266  & 0.9228  \\
			FM        &0.7702 &0.9879  &0.7853 &0.9947 &0.9177 &1.1645  &0.7035  &0.8943 \\ 
		Multiverse    &0.8667 &1.1135  &0.8290 &1.0430 &0.9316 &1.1665  &0.7275 &0.9173  \\
		  GPLVM--MF   &0.8163 &1.0144  &0.8587 &1.0841 &0.9249 &1.1715  &0.6845 &0.8758  \\
	  	   Const      &0.8322 &1.0338  &0.8991 &1.1245 &1.0077 &1.2514  &0.8290 &1.0355 \\
			\hline
			\end{tabular}
		\end{center}
	\end{table*}	

First, we show the effect of having contextual information by comparing several context-aware methods with a context-agnostic method, in here, the context-agnostic method is GPLVM--MF. We originally intended to list other context-agnostic methods, such as standard matrix factorization adopted by \cite{nguyen2014gaussian}. However, our Bayesian non-linear matrix factorization method (GPLVM--MF) outperform the ``Const" method on all datasets, while the same superiority is not guaranteed by the standard matrix factorization method reported in \cite{nguyen2014gaussian}. Thus we adopt GPLVM--MF as an appropriate context-agnostic method.

It is evident that all the context-aware methods except (1) ``Multiverse" method and (2) the Movielens 1M dataset, achieve better performance than GPLVM-FM. Since ``Multiverse" suffers from the high dimensionality of contexts on the Comoda and Sushi dataset, we use 4 out of 12 contexts for the Comoda and 3 out 7 contexts for the Sushi dataset. While there is no explicit contextual information for the Movielens-1M, the choice of model can be the most determining factor in terms of achieving a robust result. It can be even more so than incorporating contextual information. It's also noteworthy to state that by using contexts, our GPLVMF further improves performance. 
  
Then, we compare GPLVMF with other state-of-the-art context-aware methods. The performance in Table~\ref{tab:performance} shows that GPLVMF significantly outperforms other context-aware methods. Comparing with the best performance of other models, GPLVMF improves the MAE values by 4.0\%, 13.4\%, 1.2\%, and 4.7\% on the Comoda, Food, Sushi, and Movielens-1M datasets, while improves the RMSE values by 3.7\%, 12.8\%, and 4.5\% on the Comoda, Food, and Movielens-1M datasets respectively. 

Finally, we'd like to elaborate on the Sushi dataset. There are four real-valued contexts which cannot be modeled directly into GPFM model while they quantize real values into several categories \cite{nguyen2014gaussian}. In the process of quantizing, valuable information may be filtered out. However, we use these raw contextual values by fixing them as the corresponding latent variables to utilize the given data as much as possible. To confirm that the performance differences between GPLVMF and GPFM are due to directly utilizing real-valued contexts, we conduct an experiment using categorical contexts by GPLVMF, and the results are nearly the same as the GPFM, which concludes the efficiency of integrating real-valued contexts in GPLVMF.

\subsection{Optimization comparison}

  \begin{figure}[htbp]
		\begin{center}
			\includegraphics[width=0.38\textwidth]{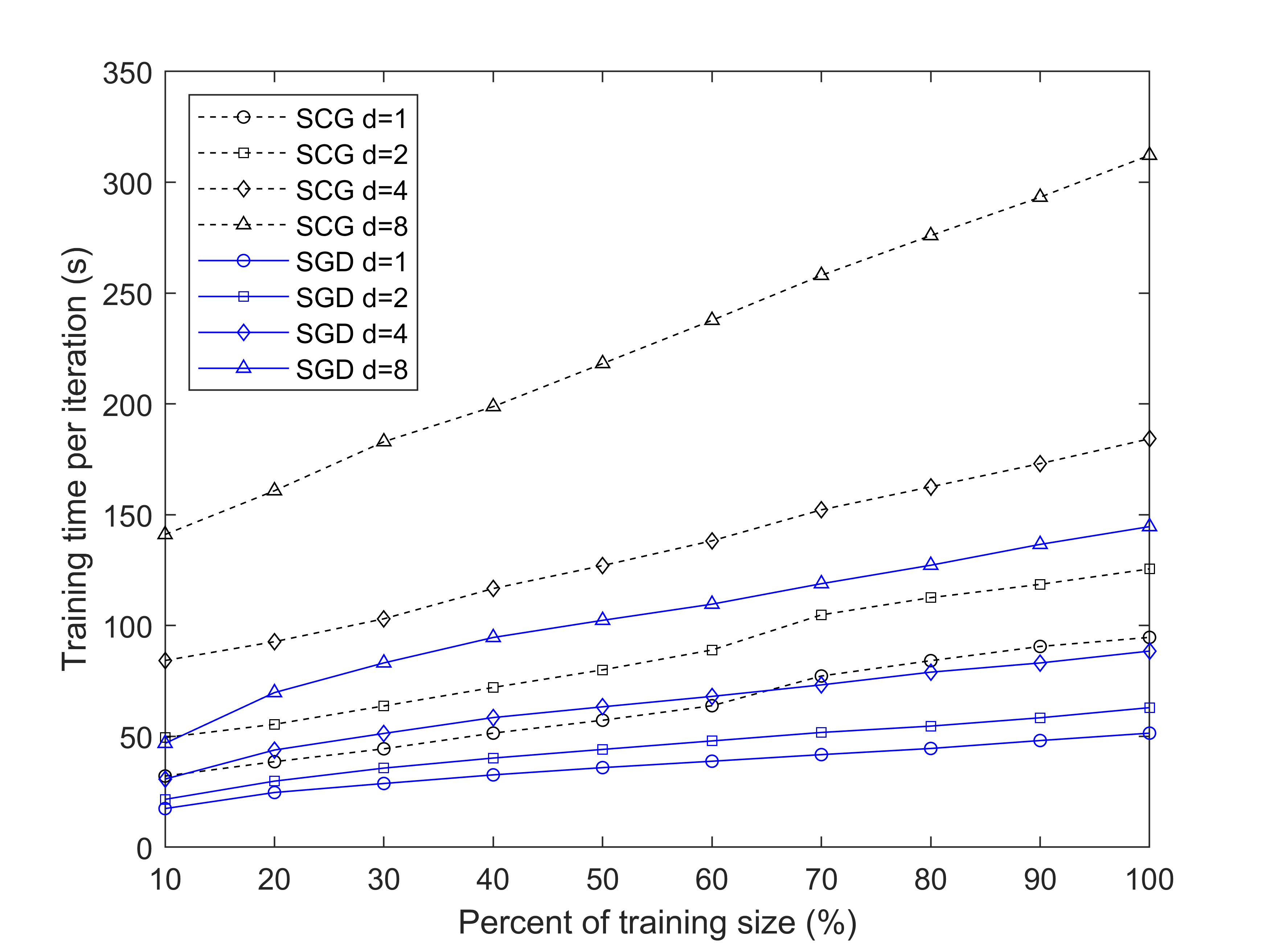}
			\caption{Training time per iteration for the Sushi dataset as a function of training size the dimensionality of latent space.\label{time}}
		\end{center}
	\end{figure}

   \begin{figure*}[htbp]
		\begin{center}
			\includegraphics[width=1.0\textwidth]{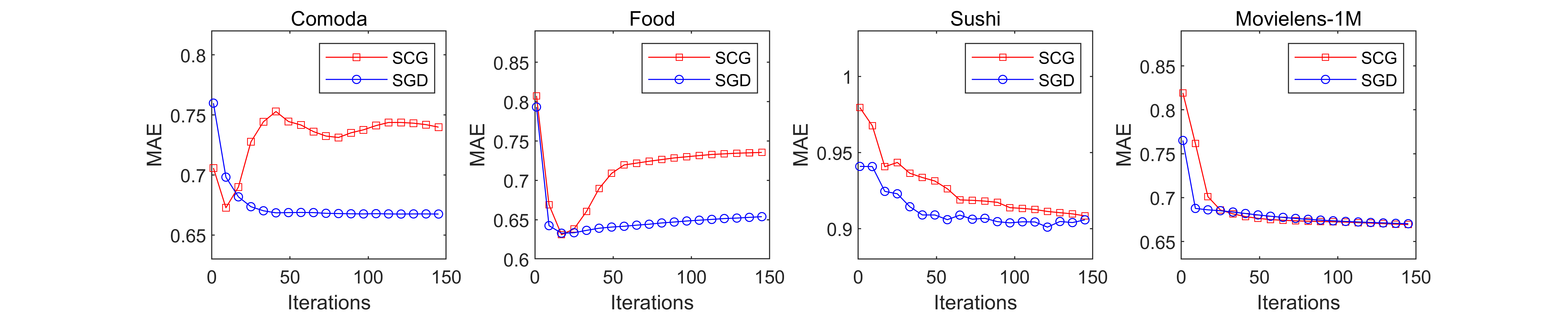}
			\caption{Comparison of the convergence curves of MAE of SGD and SCG on the four datasets.\label{convegence}}
		\end{center}
	\end{figure*}

We demonstrate the comparison of stochastic gradient descent (SGD) and scaled conjugate gradient (SCG) methods in two perspectives. First, we measure the running time for one epoch (SGD) and one iteration (SCG) against the amount of data used for training from the Sushi dataset and the dimensionality of the latent space for both SCG and SGD, where all hyperparameters are identical. We show the running time per epoch (SGD) and per iteration (SCG) in Figure \ref{time}, from which we can observe the linear growth of training time with data increasing. Besides, the number of optimization variables increases linearly with $d$. Thus the training time also scales with $d$. Overall, both SGD and SCG have the linear computational complexity for GPLVMF, while SGD is more efficient in terms of training time. 

Second, the converge curves of SGD and SCG on the four datasets are illustrated in Figure \ref{convegence}. We show that the MAE of SGD converges after about 100 epochs for all the datasets while MAE of SCG should stop at a specific iteration to obtain the best performance for the small dataset: Comoda and Food. On the other hand, the best performance between SGD and SCG has no difference. To conclude, SGD optimization has a satisfactory convergence rate and can be trained rapidly and effectively in real-world applications.

\subsection{Analysis of Contexts}

\begin{figure*}[htbp]
		\begin{center}
			\includegraphics[width=0.9\textwidth]{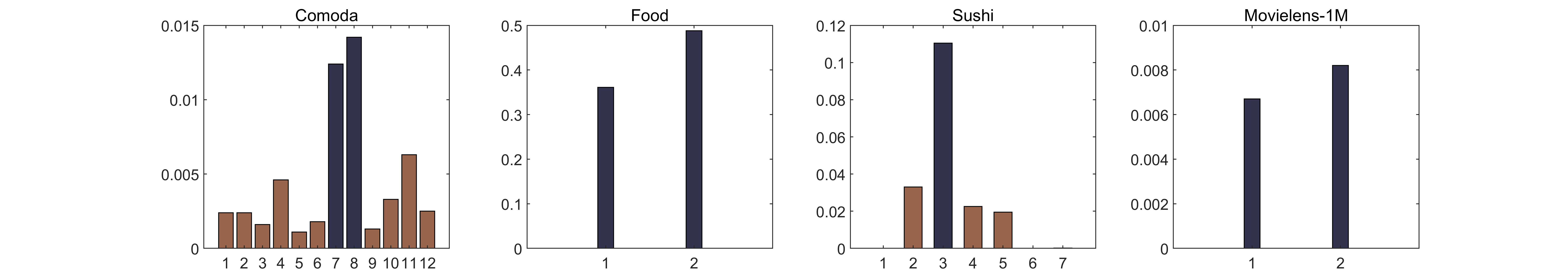}
			\caption{Inverse length-scale of the contexts on the four real datasets \label{Context}}
		\end{center}
	\end{figure*}

The ability to discover the importance of each context is essential in many regards. In this work, we seek to extract this information by performing a qualitative analysis of the {\it inverse length-scale} $\bm{\alpha}$. The original meaning of the inverse length-scale of Gaussian process latent variable model is to determine the strength of each latent dimension. For the application to CARS, the inverse length-scale can be used to automatically determine the strength of each context, as shown in Figure \ref{Context}. 

For the Comoda dataset, 2 {\it inverse length-scale} ($7$ and $8$) out of 12 contexts, are significant. These two contexts represent ``ending" and ``dominant emotions" respectively. The same conclusion has been obtained by \cite{nguyen2014gaussian} where they run experiments using one of the contexts at a time, which is computationally expensive compared to our method. On the Food dataset, the {\it inverse length-scale} of ``situation" is smaller than that of ``hunger degree", which indicates the context ``hunger degree" is more important. The same conclusion has been achieved by \cite{liu2015cot} as well. The most important context for the Sushi dataset is ``minor group". It is not too surprising since the users typically choose Sushi according to the type of it. Finally, the ``day" is less significant than ``hour" on the Movielens-1M dataset.

Finally, the impact of parameters, i.e. different number of inducing points and dimensionalities of the latent space of the item and each context, can be found in the supplementary material.

 \subsection{Impact of parameters}

\begin{figure}[htbp]
		\begin{center}
			\includegraphics[width=0.35\textwidth]{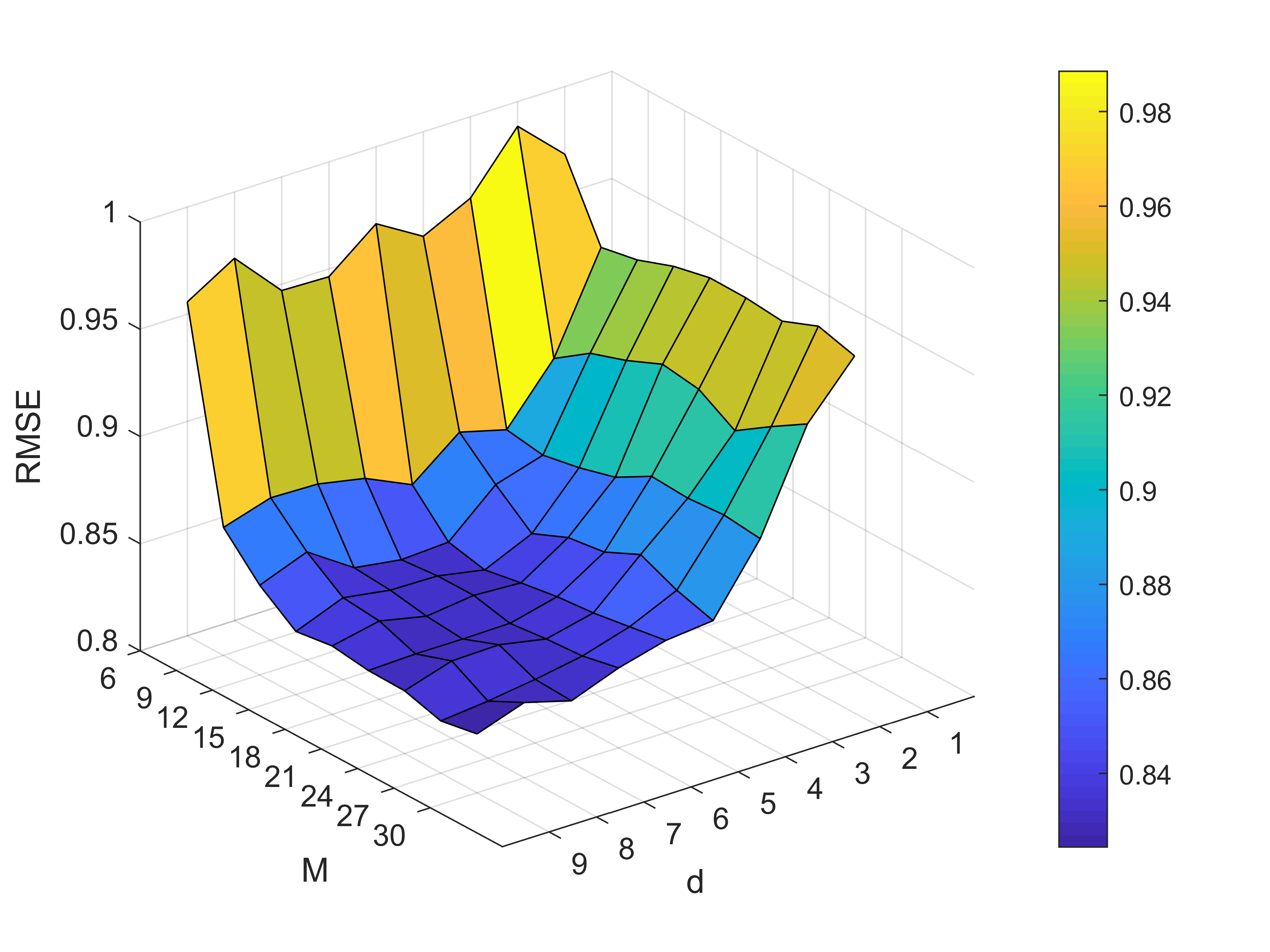}
			\caption{RMSE of GPLVMF on the Food dataset with different $d$ and $M$.\label{3d}}
		\end{center}
	\end{figure}

\begin{table}
		\caption{Empirical optimal parameters (``use mean", ``dimensionality") for each dataset.}
	   \label{tab:preference}
		\begin{center}
			\label{st}
			\begin{tabular}{|c|c|c|c|c|}
			\hline
			Dataset          & Use mean  & Dimensionality \\
			\hline
		     Comoda          &  Yes     & 1          \\                                    
         \hline
            Food            &  No      & 7        \\
			\hline
            Sushi           &  Yes     & 2        \\
			\hline
            Movielens-1M    &  No      & 6         \\
			\hline
		\end{tabular}
	\end{center}
	\end{table}	

Finally, we study the performance with using mean funtion (bias) or not, different number of inducing points $M$ and dimensionalities $d$ of the latent space of the item and each context. We first compare the performances of GPLVMF models with using mean function and different dimensionalities and list the preference to achieve the best performance in Table \ref{tab:preference}. We find that datasets with plenty of contexts (Comoda and Sushi) have a preference on the model with mean and low latent dimension while datasets with fewer contexts (Food and Movielens-1M) prefer model without the mean and high latent dimension. Then we use the Food dataset as an example to plot the value of RMSE versus $M$ and $d$, as shown in Figure \ref{3d}. With increasing $d$ and $M$, the value of RMSE decreases at first, then stays nearly stable after $d=5$ and $M=15$. For all the datasets, the parameter $M$ can be selected in a broad range, which means that the performance of GPLVMF doesn't rely on the number of inducing point very much.

\section{Conclusion}
	
In this paper, we proposed a novel Gaussian process latent variable model factorization algorithm for context-aware recommendation called GPLVMF, to effectively utilize the contextual information. Both the mean and kernel in the Gaussian process are carefully re-modeled to capture rich modeling. The experimental results of four real datasets show that GPLVMF outperforms the state-of-the-art models and can analyze the influence of each context in various datasets.

\bibliography{main}

\end{document}